\documentclass{article}

\usepackage[numbers]{natbib}

\usepackage{arxiv}

\usepackage[utf8]{inputenc} 
\usepackage[T1]{fontenc}    
\usepackage{hyperref}       
\usepackage{url}            
\usepackage{booktabs}       
\usepackage{amsfonts}       
\usepackage{nicefrac}       
\usepackage{microtype}      
\usepackage{xcolor}         

\usepackage{graphicx} 
\usepackage{multirow}
\usepackage{wrapfig,lipsum,booktabs} 
\usepackage{bbding,amsmath,amsthm,amsfonts,amssymb,amscd,pifont}
\usepackage{diagbox}

\usepackage{enumitem}

\title{DiffHand: End-to-End Hand Mesh Reconstruction via Diffusion Models}

\author{%
  Lijun Li\\
  Alibaba Group\\
  \And
  Li'an Zhuo \\
  Alibaba Group \\
  \AND
  Bang Zhang \\
  Alibaba Group \\
  \And
  Liefeng Bo \\
  Alibaba Group \\
  \And
  Chen Chen \\
  University of Central Florida \\
}

\begin{document}

\maketitle

\begin{abstract}
  Hand mesh reconstruction from the monocular image is a challenging task due to its depth ambiguity and severe occlusion, there remains a non-unique mapping between the monocular image and hand mesh.
  To address this, we develop DiffHand, the first diffusion-based framework that approaches hand mesh reconstruction as a denoising diffusion process. Our one-stage pipeline utilizes noise to model the uncertainty distribution of the intermediate hand mesh in a forward process. We reformulate the denoising diffusion process to gradually refine noisy hand mesh and then select mesh with the highest probability of being correct based on the image itself, rather than relying on 2D joints extracted beforehand. To better model the connectivity of hand vertices, we design a novel network module called the cross-modality decoder. 
  Extensive experiments on the popular benchmarks demonstrate that our method outperforms the state-of-the-art hand mesh reconstruction approaches by achieving $5.8$mm PA-MPJPE on the Freihand test set, $4.98$mm PA-MPJPE on the DexYCB test set.
\end{abstract}

\section{Introduction}
\label{sec:intro}

The field of hand mesh reconstruction has undergone rapid development in recent years. Early methods relied on depth cameras to capture hand movements and create 3D models~\citep{depthpose1, depthpose2, depthpose3, depthpose4}, but these methods were complicated and limited in accuracy. The advent of machine learning techniques, such as convolutional neural networks (CNNs), has led to significant improvements in the accuracy of hand mesh reconstruction. In recent years, deep learning models have been developed that can estimate the pose of a hand from a single RGB image, without the need for depth cameras. These models are trained on RGB hand image datasets~\citep{ho3d, freihand, dexycb, interhand26}, which enables them to learn the complex relationships between hand mesh and image features. This has led to the development of hand mesh reconstruction systems on monocular images, which have a wide range of applications in fields such as robotics, virtual reality, and human-computer interaction.

\begin{figure}[h]
	\centering
	\includegraphics[width=0.70\textwidth]{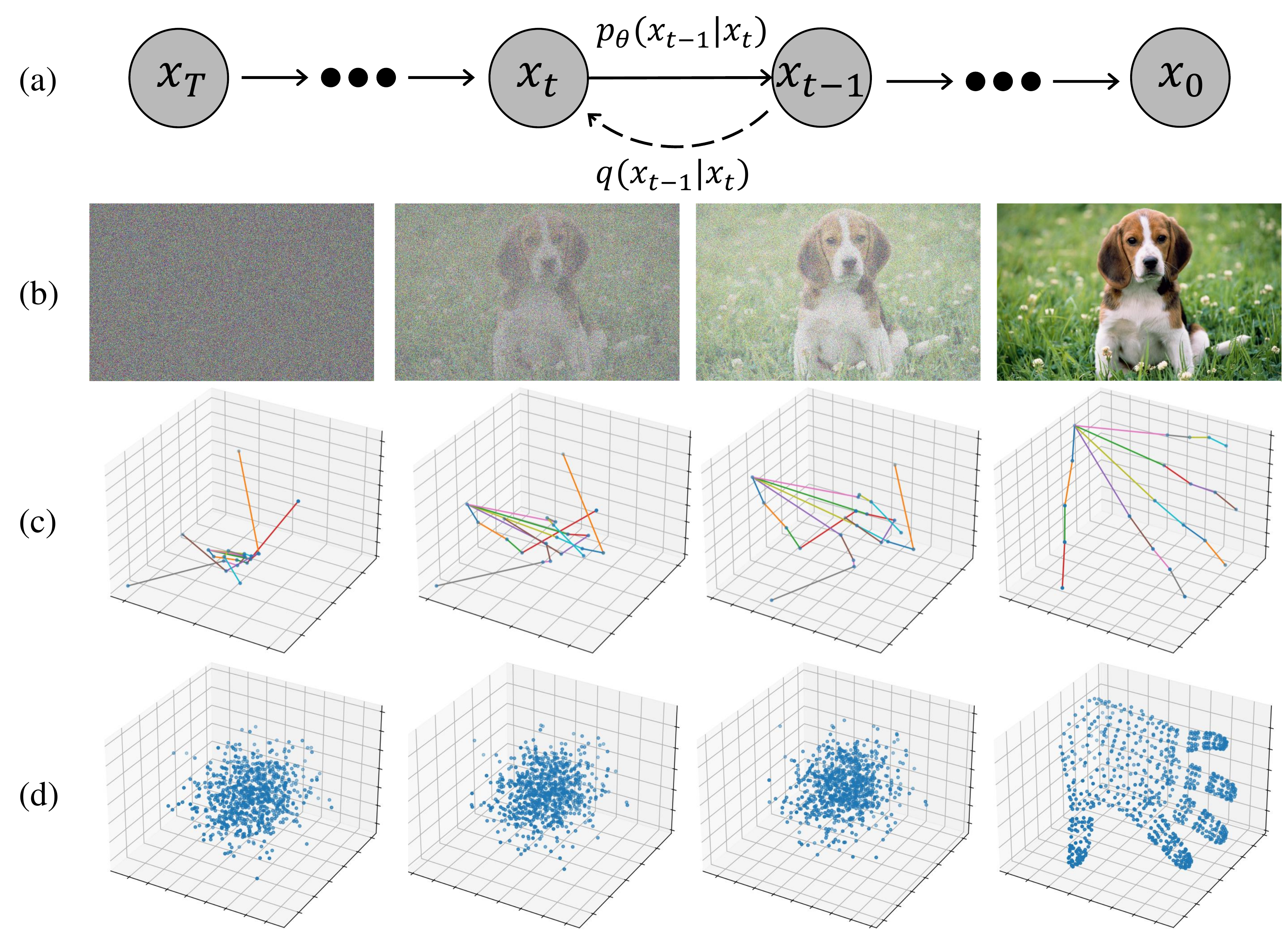}
	\caption{\textbf{Diffusion sampling process for hand mesh reconstruction.} The first row describes the sampling process from Gaussian noise to sample. The second row describes the diffusion model for image generation. The third row and fourth rows describe 3D hand joints and vertices prediction as the denoising diffusion process from noisy 3D points.}
	\label{diff_process}
\vspace{-0.3cm}
\end{figure} 


However, hand mesh reconstruction from monocular images is an ill-posed problem due to depth ambiguity and severe occlusion. The limited view and occlusion only provide information about the surface of the hand and not its internal structure, leading to a non-unique mapping between the 2D image and the 3D hand mesh. This results in suboptimal reconstruction.
As a family of probabilistic generative models, diffusion models~\citep{sohl2015deep} can address this dilemma since they can learn a data distribution and generate new samples from it. In addition to image generation~\citep{ddpm,dalle2,dreambooth}, diffusion models have demonstrated their advantages in various tasks recently, such as object detection~\citep{diffdet} and segmentation~\citep{diffseg}. However, no prior work has adapted diffusion models for hand mesh reconstruction.

For hand mesh reconstruction, diffusion models can be used to generate accurate hand mesh from noisy initialization. As shown in Figure~\ref{diff_process}, hand mesh reconstruction is analogous to image generation where the hand vertices are gradually denoised step by step. In this paper, we propose the first hand mesh reconstruction method that leverages the power of diffusion models, to learn the distribution of hand vertices given a hand image. Our method has two processes: the forward process and the reverse process. The forward process generates intermediate distributions of hand vertices for training. Specifically, given a ground truth hand vertices $V^{3D}$, the forward process gradually adds Gaussian noise to it. The intermediate noise-added result and corresponding image are used to train the model. The reverse process takes a set of random points initialized from a Gaussian distribution and denoises them into 3D hand vertices conditioned on the image feature. The reverse process is crucial for both training and inference.

Although there are some works on 3D diffusion-based human pose estimation, such as~\citet{diffpose2d},~\citet{diffpose_gcn} and~\citet{diffpose_gcn1}, they are all two-stage models. These models first generate 2D human joints from an image and then predict 3D joints given 2D keypoints via the sampling process of Denoising Diffusion Probabilistic Models (DDPM)~\citep{ddpm}. Graph convolution networks or self-attention layers are used to model the connections between joints during the sampling process. However, these approaches have some drawbacks. For instance, the pose estimation accuracy heavily relies on the performance of the person keypoint detector, which can lead to inferior performance in complex scenarios. Moreover, the use of an isolated detector increases the computational cost.

In contrast, our method is superior in several ways. Firstly, we introduce the cross-modality decoder to better utilize vertex connectivity. This approach can largely improve reconstruction accuracy when compared to vanilla self-attention. Second, our framework can be trained end-to-end without the need for an off-the-shelf detector and therefore can avoid errors from the off-the-shelf detectors. According to~\citep{diffpose2d}, incorporating an image instead of 2d human joints into diffusion-based pose estimation methods mostly leads to degraded performance. In our paper, we successfully fuse images into mesh reconstruction and achieve state-of-the-art (SOTA) results on hand mesh datasets. Our contributions are summarized as follows:
\begin{itemize}[leftmargin=*,itemsep=2pt,topsep=0pt,parsep=0pt]
\item We propose a novel one-stage 3D hand mesh reconstruction framework that is trained end-to-end and generates hand meshes from hand images. Our model is customizable and can accept more inputs, for example, depth image, to further improve the reconstruction accuracy.
\item The cross-modality decoder is designed to better explore the hand vertex connectivity and utilize image features by designing U-shaped vertex block to gain the rich multi-scale vertex features and attention block to fuse global embedding and image features.
\item Compared with other hand mesh reconstruction works, our method achieves state-of-the-art (SOTA) performance. Despite designing specific modules to estimate hand mesh in hand-object interaction, our method can still achieve SOTA results in two hand-object interaction datasets, DexYCB~\citep{dexycb} and HO-3D V2~\citep{ho3d}.
\end{itemize}

\section{Related Work}
\subsection{3D Hand Pose Estimation}
Benefiting from the development of the parametric models (e.g., MANO~\citep{romero2017embodied}), 
the framework of 3D hand pose estimation generally consists of two parts, i.e., a feature extractor achieved by the deep networks, and a regressor designed to predict the coefficients of the parametric model~\citep{boukhayma20193d, moon2020i2l, tang2021towards} or the vertices of the hand mesh directly~\citep{lin2021end, graphormer}.
Previous works mainly differ in the architecture of the regressor for better reconstruction.
For example, 
\citet{boukhayma20193d} proposes the first end-to-end deep learning-based method that adjusts the final fully connected layer to output the parametric coefficients.
\citet{moon2020i2l} uses pixel-based 1D heatmap to preserve the spatial relationship in
the input image and models the uncertainty of the prediction to obtain dense mesh vertex localization.
\citet{lin2021end} introduces a multi-layer transformer regressor to model both vertex-vertex and vertex-joint interactions for better 3D hand reconstruction.
\citet{graphormer} presents a graph convolution-based transformer to combine graph convolution and self-attention to model both local and global interactions for hand mesh reconstruction.
\citet{tang2021towards} designs multi-task decoders and decouples the hand-mesh reconstruction task into multiple stages to ensure robust reconstruction.

\subsection{Diffusion Model}
As a family of deep generative models, diffusion models~\citep{sohl2015deep} begin with adding random noise into the raw data gradually, and then regard the generation process as a gradual denoising process. These models have demonstrated remarkable contributions in numerous areas of computer vision, i.e., 
image colorization~\citep{song2020score}, 
super-resolution~\citep{saharia2022image, li2022srdiff},
semantic segmentation~\citep{baranchuk2021label, brempong2022denoising}, 
object detection~\citep{diffdet}.
For instance, 
~\citet{li2022srdiff} proposes a novel single-image super-resolution diffusion probabilistic model to generate diverse and realistic super-resolution predictions.
DiffusionDet~\citep{diffdet} applies diffusion models to object detection to recover bounding boxes from noise. 
~\citet{baranchuk2021label} demonstrates
that diffusion models can also serve as an instrument for semantic segmentation.
Recently, ~\citet{diffpose_gcn} and ~\citet{diffpose_gcn1} apply the diffusion model to the 3D human pose estimation, which adds noise to the 3D human joints and produces the predictions in the denoising process. 

In this paper, we utilize the diffusion model to perform hand mesh reconstruction. To our best knowledge, this is the first attempt to introduce the diffusion model into hand mesh reconstruction.

\begin{figure}[h]
	\centering
	\includegraphics[width=0.875\textwidth]{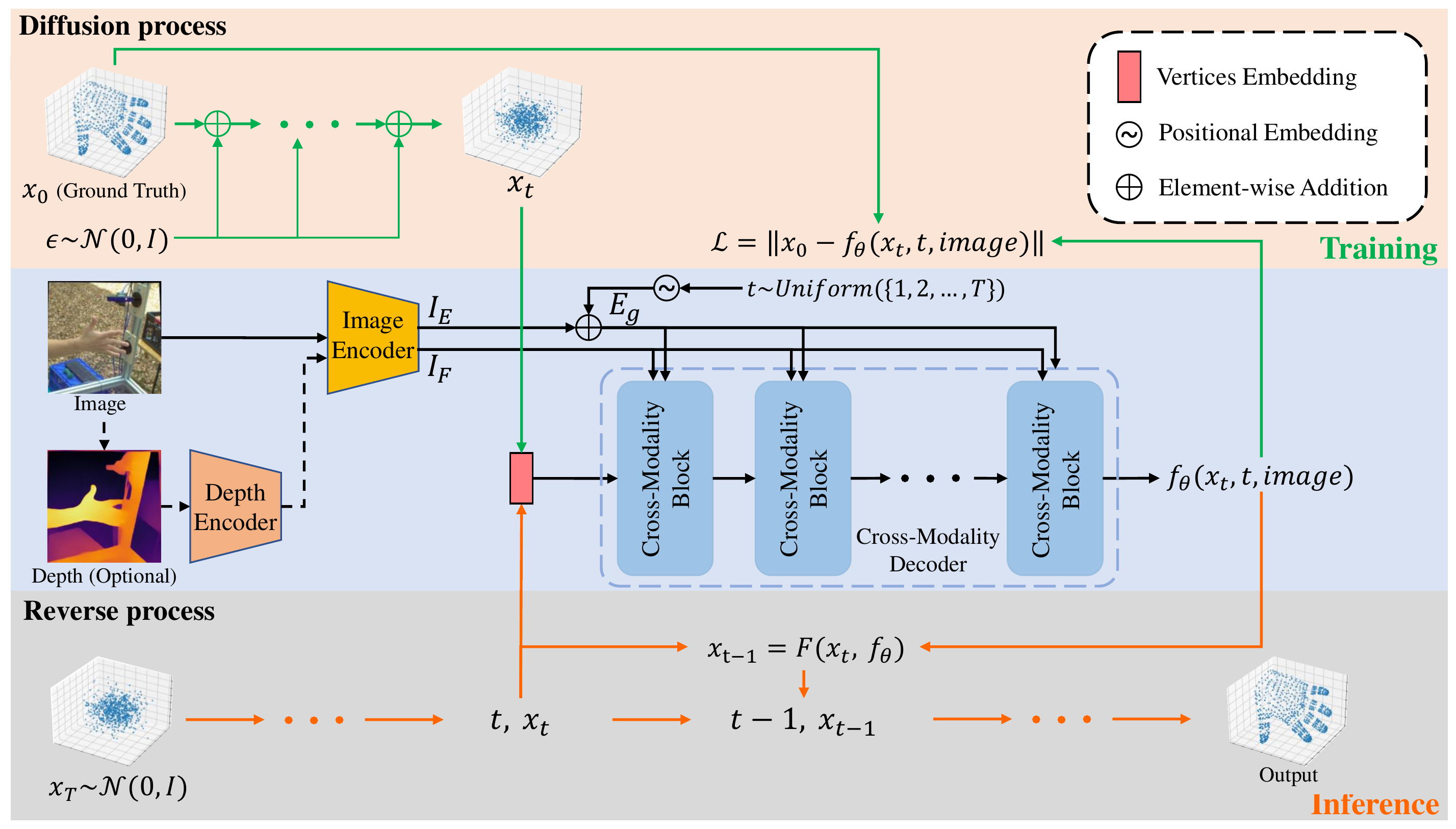}
	\caption{\textbf{Overview of our network}. During training, each diffused intermediate vertices are utilized as input to the network and denoised to predict the ground truth conditioned on image feature. Ground truth and prediction formulate the loss function to update parameters. During inference, randomly initialized points sampled from Gaussian distribution are fed to our denoiser network $\mathcal{D}$ along with the image feature generated from image encoder $\mathcal{E}$. At each timestep $t$, DiffHand predicts a clean sample $\hat{x_0}$ and diffuses it back to $x_{t-1}$ according to Equation~\ref{get_xt-1}. After multiple denoising steps, the final hand vertices $\hat{x_0}$ are generated.
	}
	\label{pipeline}
\vspace{-0.3cm}
\end{figure} 

\section{DiffHand}
\label{sec:method}
Given a hand image $I$, hand mesh reconstruction is to predict 3D positions of hand joints $J^{3D}\in{R^{M\times3}}$ and vertices $V^{3D}\in{R^{N\times3}}$, $M$ and $N$ are number of joints and vertices respectively. Hand vertices can be treated as a set of points with predefined topology. Inspired by diffusion-based generative models~\citep{stablediffusion}, a randomly initial distribution can be transformed into a deterministic output under additional information. We aim to solve the reconstruction task via the diffusion model that recovers hand vertices from random initialization given image features. With the help of DDPM, our method can solve the uncertainty of non-unique mapping between the 2D image and the 3D hand mesh to some extent. Under different conditions $c$, our network models the distribution $p(x\mid{c})$ which is used to predict $x_0$ from noisy inputs $x_t$. Remarkably, our framework is a generic architecture that enables multiple forms of condition $c$ including RGB images, hand silhouettes, 2D hand keypoints, and depth images.

\subsection{Diffusion Model}
\textbf{Preliminary.} Diffusion model is a probabilistic generative model that consists of two parts: forward process and reverse process. The forward process gradually adds Gaussian noises of predefined mean and variance to the original data to transform it into Gaussian distribution in the end. Given a set of variance schedule $\{{\beta_t}\}_{1}^{T}\in(0,1)$, the forward process is defined as \citep{ddpm}:
 \begin{eqnarray}
 q(x_t{\mid}x_0) := \prod_{t=1}^{T}{q(x_t{\mid}x_{t-1})}\\
 q(x_t{\mid}x_{t-1}):=\mathcal{N}(x_t;\sqrt{1-\beta_t}x_{t-1}, \beta_t\mathbb{I}),
 \end{eqnarray}
Afterwards, we can directly obtain a noisy sample ${x_t}$ at an arbitrary timestep $t$ from the data distribution:
\begin{eqnarray}
    q(x_t{\mid}x_0):=\mathcal{N}(x_t;\sqrt{\bar{\alpha_t}}x_0, (1-\bar{\alpha_t})\mathbb{I}),
    \label{x_forward}
\end{eqnarray}
where $\bar{\alpha_t}:={\prod_{t=1}^{T}}{\alpha_t}$ and $\alpha_t:=1-{\beta_t}$. When $\alpha_t$ is small enough, the noisy sample $x_t\sim{\mathcal{N}(0,1)}$.

Since calculating $q(x_{t-1}\mid{x_t})$ depends on the entire data distribution, as proposed by ~\citet{ddpm}, we can approximate $q(x_{t-1}\mid{x_t})$ with a neural network $\theta$. The reverse process iteratively removes noise by modeling a neural network $\theta$ to get the posterior probability: 
\begin{eqnarray}
    p_\theta(x_{t-1}\mid{x_t})=\mathcal{N}(x_{t-1};\mu_\theta(x_t,t),\Sigma_{\theta}(x_t,t)).
    \label{get_xt-1}
\end{eqnarray}
where $\sigma$ is defined by $\beta_t$. DDPM estimates the noise $\epsilon_\theta(x_t,t)$ between consecutive samples $x_{t-1}$ and $x_t$ and the training loss function is defined as:
\begin{eqnarray}
    \mathcal{L}=\|{\epsilon - \epsilon_\theta(x_t,t)}\|^2.
\end{eqnarray}
More details can be seen in DDPM~\citep{ddpm}.

\subsection{Forward and Reverse Reconstruction Processes}
In this work, we solve the hand mesh reconstruction with diffusion model. Data samples are a set of points or, in other words, hand vertices. Given a random noise and an image, the noise is progressively denoised into hand mesh corresponding to the image. In order to learn the correlation between the image and hand vertices, we require the intermediate vertices and their ground truth to train the network. The intermediate samples are obtained from the forward process. We denote the hand vertices $v$ as $x$ in the following.

\textbf{Forward process.} As shown in Figure~\ref{diff_process}, given a ground truth hand vertices $x_0\in{R^{N\times3}}$, Gaussian noises of increasing variance are added gradually. Sampling from timestep $t\sim{U(0,T)}$, we generate intermediate vertices $x_t\in{R^{N\times3}}$ by adding Gaussian noise indicated in Equation~\ref{x_forward}. With the intermediate supervisory signals, it is allowed to optimize the network to perform denoising.  Instead of predicting noise $\epsilon_t$ in ~\citep{ddpm}, we follow DALL-E 2~\citep{dalle2} to directly predict the signal itself that leads to better results, with the following objective:
\begin{eqnarray}
    \mathcal{L}=\parallel{f_\theta(x_t,t,I) - x_0}\parallel{^2}.
    \label{mse_loss}
\end{eqnarray}

\textbf{Reverse process.} In the reverse process, $x_T$ is drawn from Gaussian distribution, and the image is fed to image encoder $\mathcal{E}$ to get image embedding $I_E$ and image feature map $I_F$ as condition. For each sampling step, the estimated hand vertices from the last sampling step, the image features, and the timestep are sent to cross-modality decoder to predict the newer hand vertices. During the sampling step, DDIM~\citep{ddim} is adopted to estimate the intermediate vertices for the next step.

\subsection{Network Architecture}
Different from the traditional diffusion-based models such as Stable Diffusion~\citep{stablediffusion} and GLIDE~\citep{glide}, which use UNet~\citep{unet} as the main network for image generation, we design a different network that is suitable for hand mesh reconstruction. As shown in Figure~\ref{pipeline}, our framework consists of two parts: an image encoder that extracts image features from the image and a cross-modality decoder that takes image features as condition and progressively refines the hand vertices.
The cross-modality decoder is proposed to explore the structured relations between hand vertices and adaptively attend to the most relevant features across different scales of vertices. We design a U-shaped vertex block to model the relation between hand vertices, and an attention block to enhance point features with global embedding and image feature map.

\textbf{Image encoder.} We directly use a standard image classification backbone as an image encoder to extract the image features. CNN-based network, HRNet~\citep{hrnet}, and transformer-based network, Deit~\citep{deit}, are used as the image encoder. From the function following, we can get the image feature map $I_F$  and image embedding $I_E$ from the image encoder: $I_F,\ I_E=\mathcal{E}(I)$. The image feature map is a high-dimensional representation of the input image that captures the spatial relationships between the pixels. Image embedding is a lower-dimensional vector representation of the image feature map that captures the semantic content of the image.

\begin{figure}[h]
	\centering
	\includegraphics[width=0.875\textwidth]{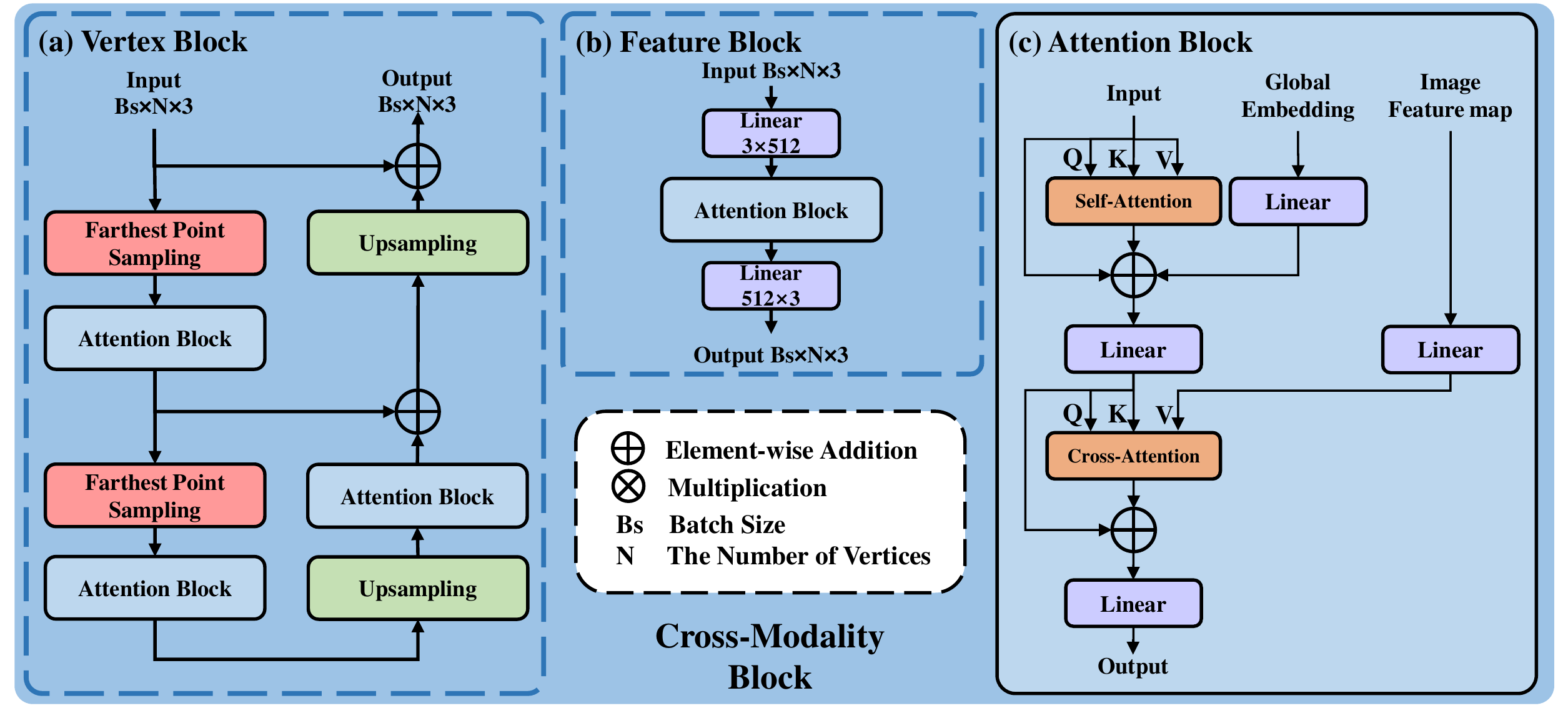}
	\caption{\textbf{Detailed components of our cross-modality block}. There are 3 cross-modality blocks in cross-modality decoder to model the hand vertices structure. Each cross-modality block is made up of one vertex block and one feature block. (a) shows the vertex block is made up of two downsampling modules and two upsampling module. (b) denotes the feature block after each vertex block. It models the correlation in feature space. (c) is the architecture of attention block. Attention block is made up of two sub-attention modules, self-attention block and cross-attention block. In self-attention block, point feature and global embedding are fused. In cross-attention block, Image feature maps are injected into points to generate points conditioned on the given image feature map.
	}
	\label{point_block}
\vspace{-0.4cm}
\end{figure}

\textbf{Cross-modality decoder.} As neural network $f_\theta(x_t, t, I)$ is crucial for denoising, we introduce our cross-modality decoder in the following. Since the level of noise is denoted by the timestep $t$, it needs to provide the timestep to cross-modality decoder. As shown in Figure~\ref{pipeline}, the cross-modality decoder takes image feature map $I_F$, intermediate vertices embedding $x_t$, and global embedding $E_g$ as inputs.
Our cross-modality decoder has 3 cross-modality blocks that progressively perform cross-attention with downsampled vertices and image features. 
Each cross-modality block has a vertex block followed by a feature block. 
In order to better model the structure of hand vertices, we design a U-shaped module called vertex block with lateral residual connection since different level represents varying degrees of coarseness to fineness in hand structures, as shown in Figure~\ref{point_block} (a). The vertex block consists of downsampling and upsampling parts. One of the purposes of farthest point sampling is to reduce the point dimension. For each downsampling, the downsampling rate is the same and the point cardinality is reduced from $N$ to $N/4$. For upsampling, a multilayer perceptron (MLP) is used to increase the point cardinality. After each downsampling, our attention block is applied to combine the information of the point feature, global embedding, and image feature. In each attention block shown in Figure~\ref{point_block} (c), there are self-attention block and cross-attention block. After self-attention, we enhance output feature by adding global embedding $E_g$. By utilizing Equation~\ref{att}, we can get output $Y$ of self-attention block, where $W_Q$, $W_K$ and $W_V$ are linear projections to generate $Q$, $K$ and $V$ , $d$ is the dimension of $X$. Cross-attention block further incorporates point feature with the image feature maps,
 enabling each vertex adaptively attend to the most relevant features.
The feature block is then performed after the vertex block as shown in Figure~\ref{point_block}(b). Vertex block and feature block enhance point features by modeling correlation in point space and feature space respectively.
\begin{eqnarray}
    Q=W_QX, K=W_KX, V=W_VX\\
    Y=softmax(\frac{QK^T}{\sqrt{d}})V+Linear(E_g)
    \label{att}
\end{eqnarray}

\textbf{Extra condition.} 
Our model is capable of incorporating additional inputs to guide the generation of hand vertices. For example, we can use a depth image as an extra condition as shown in Figure~\ref{pipeline}. Depth estimation method Midas~\citep{midas} is utilized to obtain the corresponding depth image from the hand image. To better preserve the ability of the original network and enhance it by adding new conditions, we first train the original network. After training the multi-modality encoder, we finetune it with an extra block which is fed with depth image for a few epochs. This block is composed of several convolutional layers, all of which are initialized with zero weights and biases. At last, we can get the vertices prediction by $\hat{x_0}=f_\theta(x_t, t, I, depth)$.
\subsection{Training and Inference}
\textbf{Training.} As shown in Figure~\ref{pipeline}, the forward process generates intermediate corrupted vertices $x_t$ from ground truth vertices $x_0$ and Gaussian noise defined by cosine noise variance schedule $\beta$ and timestep $t$. For each timestep, we use a sinusoidal function to translate timestep $t$ to timestep embedding $E_t$. Then image embedding $I_E$ and timestep embedding $E_t$ are summed to form the global embedding $E_g$. The vertices embedding $x_t$, global embedding $E_g$, and image feature map $I_F$ are all adopted as inputs to our cross-modality decoder. Afterward, the decoder result of vertices prediction $\hat{x_0}$ is generated. The network is supervised by MSE loss in Equation~\ref{mse_loss}. Additionally, traditional training losses for hand mesh reconstruction are also applied, including MSE loss for 3D hand vertices $V^{3D}$, joints $J^{3D}$, 2d projected joints $J^{2D}$. To project the 3D joints into 2D space, we utilize the estimated camera parameters. The vertex and joint losses are defined as follows:
\begin{eqnarray}
    V^{3D}=f_\theta(x_t,t,I),J^{3D}=WV^{3D},\\
    \mathcal{L}_{vertex}=\sum_{i=0}^{N-1}{\|V^{3D}_{i}-V_{i}^{3D,GT}}\|^2,\\
    \mathcal{L}_{joint}=\sum_{i=0}^{M-1}\sum_{d\in\{3D, 2D\}}{\|J_{i}^{d}-J_{i}^{d,GT}}\|^2,
    \label{3d2djoint_loss}
\end{eqnarray}
where $M$, $N$ represents the number of joints and vertices, $W\in R^{M\times N}$ is the weight matrix that sample the expected joints from the vertices, and $d$ denotes whether the computation is for 3D or 2D. To guarantee the geometric continuity of the predicted vertices, smoothness loss is applied which regularizes the consistency of the normal direction between the predicted and the ground truth mesh:
\begin{eqnarray}
\mathcal{L}_{smooth}=\sum_{f=0}^{F-1}\sum_{j=0}^{2}{\|e_{f,j}\cdot{n_{f}^{GT}}\|_1},
\end{eqnarray}
where $f$ means the face index of hand mesh, $j$ means the edge of face $f$ and $n^{GT}$is the ground truth normal vector of this face. Overall, our training loss is the combination of all the loss
\begin{eqnarray}
    \mathcal{L}=\mathcal{L}_{vertex}+\lambda_{joint}\mathcal{L}_{joint}+\lambda_{smooth}\mathcal{L}_{smooth},
    \label{overall_loss}
\end{eqnarray}
where $\lambda_{joint}$ and $\lambda_{smooth}$ are the hyperparameters to balance these items.

\textbf{Inference.} Our diffusion model progressively denoises noisy inputs to generate hand vertices. At first, initial hand vertices $x_t$ are sampled from the Normal distribution. Similar to the training process, $x_t$, global embedding $E_g$, and image feature map $I_F$ are all adopted as inputs to our cross-modality decoder, in every timestep $t$. Then we predict the clean vertices $\hat{x_0}$ and noise it back to $x_{t-1}$ by calculating $q(x_{t-1}\mid{x_t, \hat{x_0}})$. The inference procedure will be iterated $T$ times until the final clean vertices are achieved. Note that the inference steps $T$ is an adjustable parameter that controls the trade-off between accuracy and efficiency.

\section{Experiments}
\label{sec:exp}

\subsection{Experiment Setup}
\textbf{Dataset.} We conduct our experiments on three widely used datasets for hand mesh reconstruction. Freihand~\citep{freihand} is a single-hand dataset with various views and backgrounds. The training set contains 130K images and the test set contains 4K images. HO-3D V2~\citep{ho3d} is a hand-object interaction dataset with occlusion. There are 66K images for training and 11K images for testing. DexYCB~\citep{dexycb} is a large hand grasping of object dataset. The dataset contains 582K images of 10 subjects on 20 objects from 8 views. We use the default S0 split for training and testing. For a fair comparison, we follow the setup of ~\citet{handoccnet} and filter invalid samples in the official split. Severe occlusions introduced by the objects make DexYCB dataset more challenging for hand mesh recovery. All datasets provide MANO parameters for each image.

\textbf{Evaluation metrics.}
To evaluate these methods, we report results by four standard metrics: Mean Per Joint Position Error ($E_J$), the Mean Per Joint Position Error with Procrustes Alignment ($E_{PJ}$),   Mean Per Vertex Position Error ($E_V$), the Mean Per Vertex Position Error with Procrustes Alignment ($E_{PJ}$) in millimeters (mm). As our task is focused on mesh recovery, the $E_{V}$ and $E_{PV}$ are given priority consideration in the ablation study.

\begin{wraptable}{r}{6.5cm}
\vspace{-18pt}
    \caption{Comparing with state-of-the-art on Freihand test set ($E_{PV}$/$E_{PJ}$(mm)$\downarrow$). The best results are highlighted in \textcolor{red}{red}, second best is highlighted in \textcolor{blue}{blue}. Non-available results are marked with ”-”
    }
    \begin{tabular}{c|cc}
       \toprule
	   \diagbox{method}{metric} & $E_{PJ}$$\downarrow$ & $E_{PV}$$\downarrow$ \\
       \midrule
       YoutubeHand~\cite{youtubehand} & 8.4 & 8.6 \\
       I2uv-handnet~\cite{i2uv} & 6.9 & 6.7  \\
       I2L-MeshNet~\cite{moon2020i2l} & 6.7 & 6.9 \\
       METRO~\cite{lin2021end} & 6.7 & 6.8 \\
       FastVIT~\cite{fastvit} & 6.6 & 6.7 \\
       Mobrecon~\cite{mobrecon} & 6.1& 6.2 \\
       MeshGraphormer~\cite{graphormer} & \textcolor{blue}{5.9} & \textcolor{red}{6.0} \\
       \midrule
       DiffHand(Ours) & \textcolor{red}{5.8} & \textcolor{blue}{6.1}\\
       \bottomrule
    \end{tabular}
    \label{freihand_comp}
    \vspace{-0.3cm}
\end{wraptable}

\textbf{Implementation details.} The input images are resized to $224\times224$ for the Freihand dataset and $256\times256$ for the other datasets. HRNet-W64~\citep{hrnet} and Deit-Large~\citep{deit} are selected as the image encoders. For the cross-modality decoder, we apply three cross-modality blocks and each is made up of one vertex block and one feature block. For all experiments, the implementations are done using Pytorch~\citep{pytorch}. We train all models with hand images using AdamW optimizer for 100 epochs. The hyperparameters $\lambda_{joint}$ and $\lambda_{smooth}$ are set to 1 and 0.05 respectively. The initial learning rate is $1e^{-4}$ and the batch size is 64. All experiments are performed on 4 NVIDIA Ampere A100 GPUs. For depth estimation, Midas method~\citep{midas} is used. As to DDIM sampling, we set the inference step to 10. 




\subsection{Comparison with State-of-the-Art Methods}
 Our method is compared with state-of-the-art hand mesh reconstruction methods on three datasets, Freihand~\cite{freihand}, HO-3D V2~\cite{ho3d}, and DexYCB~\cite{dexycb}. For Freihand and HO-3D V2 dataset, we use Deit~\cite{deit} as image encoder. For DexYCB dataset, we use HRNet~\cite{hrnet} as image encoder. The results are presented in Table~\ref{freihand_comp}, Table~\ref{dexycb_comp} and Table~\ref{ho3d_comp}. On Freihand dataset, we perform test-time augmentation like MeshGraphormer~\cite{graphormer}. As shown in Table~\ref{freihand_comp}, our method outperforms SOTA method MeshGraphormer and other six competitive methods over $E_{PJ}$. On the two hand-object interaction datasets, although there is no special design for modeling hand-object interaction, our method still performs the best on $E_{PJ}$ and $E_J$ where most works report performance. DexYCB dataset is more challenging than Freihand dataset due to severe object occlusion. On DexYCB dataset, it can be seen that we surpass the other by a clear margin both on $E_{PJ}$ by 0.8mm lower and $E_J$ by 1.9mm lower. There are almost 15\% lower on both metrics. On HO-3D V2 dataset, MutualAttention~\cite{mutualatt} explicitly predicts object meshes to help hand mesh reconstruction. Comparing with the SOTA method MutualAttention~\cite{mutualatt}, we obtain similar performance on $E_J$ but much lower on $E_{PJ}$ without using the object information. For the other SOTA method HandOccNet~\cite{handoccnet}, we also obtain similar performance on $E_{PJ}$, but 1.2mm lower error on $E_J$. Though our method is the first diffusion-based framework in hand mesh reconstruction, we can outperform other SOTA methods.
 
\begin{figure}[!t]
	\centering
	\includegraphics[width=0.8\textwidth]{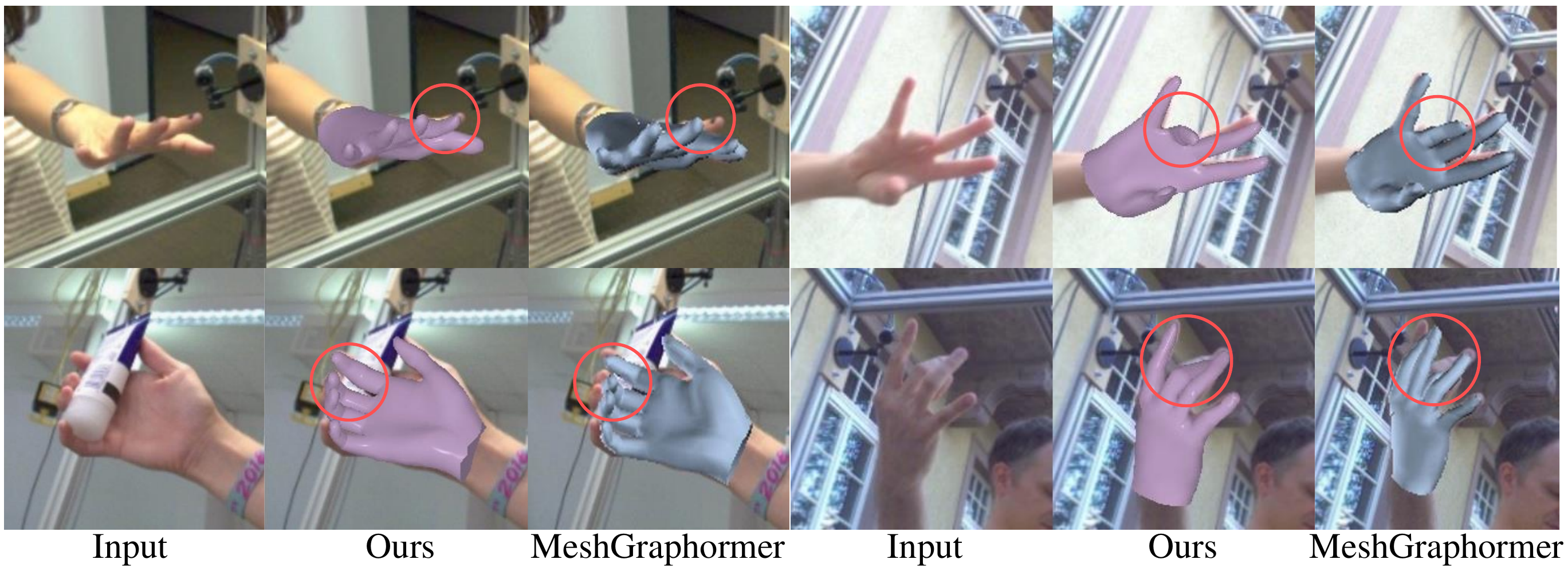}
	\caption{Comparison of qualitative results between our method and MeshGraphormer~\citep{graphormer} on Freihand test set. Red circles highlight positions where our methods generate better results.}
	\label{comparision}
\vspace{-0.3cm}
\end{figure}

\textbf{Qualitative results.} Some qualitative comparisons with MeshGraphormer~\citep{graphormer} on the Freihand test set are shown in Figure~\ref{comparision}. It can be seen that our method produces a more accurate hand mesh that aligns well with the given hand image. Our method is also less sensitive to special gestures and challenging views. More qualitative results can be seen in supplementary material.

\begin{table}[!t]
    \begin{minipage}[c]{.45\linewidth}
          \caption{Comparing with state-of-the-art on DexYCB test set ($E_{J}$/$E_{PJ}$(mm)$\downarrow$). The best results are highlighted in \textcolor{red}{red}, second best is highlighted in \textcolor{blue}{blue}. Non-available results are marked with ”-”}
        \centering
        \small
        \begin{tabular}{c|cc}
   \toprule
   \diagbox{method}{metric} & $E_{J}$$\downarrow$ & $E_{PJ}$$\downarrow$ \\
   \midrule
   Homan~\cite{homan} & 18.88 & - \\
   IdHandMesh~\cite{idhandmesh} & 16.63& - \\ 
   AlignSDF~\cite{alignsdf} & 15.7 & - \\
   ~\citet{tsecvpr22} & 15.3 & -  \\
   ~\citet{liusemi} & 15.28& 6.58 \\
   gSDF~\cite{gsdf} & 14.4 &  - \\
   HandOccNet~\cite{handoccnet} & \textcolor{blue}{14.04} & \textcolor{blue}{5.8} \\
   \midrule
   DiffHand(Ours) & \textcolor{red}{12.1} & \textcolor{red}{4.98} \\
   \bottomrule
\end{tabular}
\label{dexycb_comp}
        \end{minipage}
        \hspace*{2em}%
    \begin{minipage}[c]{.45\linewidth}
      \centering
      \small
        \caption{Comparing with state-of-the-art on HO-3D V2 test set ($E_{J}$/$E_{PJ}$(mm)$\downarrow$). The best results are highlighted in \textcolor{red}{red}, second best is highlighted in \textcolor{blue}{blue}. Non-available results are marked with ”-”}
        \begin{tabular}{c|cc}
   \toprule
   \diagbox{method}{metric} & $E_{J}$$\downarrow$ &  $E_{PJ}$$\downarrow$ \\
       \midrule
       Homan~\cite{homan} & 26.8 & -  \\ 
       Keypoint Trans.~\cite{keypointtrans} & 25.7 & -  \\ 
       Artiboost~\cite{artiboost} & 25.3 & -  \\ 
       ~\citet{liusemi} & - & 10.2  \\
       Mobrecon~\cite{mobrecon} & - & \textcolor{blue}{9.2}  \\
       HandOccNet~\cite{handoccnet} & 24.9 &  \textcolor{red}{9.1}  \\
       MutualAttention~\cite{mutualatt} & \textcolor{blue}{23.8} & 10.1  \\ 
       \midrule
       DiffHand(Ours) & \textcolor{red}{23.7} & 9.3 \\
       \bottomrule
    \end{tabular}
    \label{ho3d_comp}
    \end{minipage} 
\vspace{-0.3cm}
\end{table}

\begin{table}
\centering
\caption{Effect of cross-modality decoder and diffusion probabilistic model on Freihand test set ($E_{PV}$/$E_{V}$/$E_{PJ}$/$E_J$(mm)$\downarrow$). It is shown that both diffusion and cross-modality decoder help to reduce the error by a clear margin.
}
\begin{tabular}{cc|cccc}
   \toprule
cross-modality decoder & diffusion & $E_{PV}$$\downarrow$ & $E_{V}$$\downarrow$ & $E_{PJ}$$\downarrow$ & $E_J$$\downarrow$ \\
   \midrule
   \ding{55} & \ding{55} & 7.87 & 15.49 & 7.08 & 14.98 \\
   \ding{55} & \ding{51} & 6.99 & 13.93 & 6.74 & 13.99 \\
   \ding{51} & \ding{55} & 7.09 & 13.94 & 6.59 & 13.51 \\
   \ding{51} & \ding{51} & \textbf{6.48} & \textbf{13.16} & \textbf{6.29} & \textbf{13.34} \\
   \bottomrule
\end{tabular}
\label{ablation_net}
\vspace{-0.3cm}
\end{table}

\subsection{Ablation Study}
\textbf{Network variants.} In order to justify the effectiveness of our diffusion probabilistic model and cross-modality decoder, we construct different variants with the alternative usage of diffusion and cross-modality decoder. If the diffusion is not used, only the image is fed as input and the architecture is similar to the middle of Figure~\ref{pipeline} without forward and reverse process. The networks without diffusion take hand images as input and directly decode hand vertices from images without iterative refinement. If the cross-modality decoder is not used, it is replaced with a vanilla self-attention module which performs  self-attention after the summation of global embedding $E_g$ and intermediate vertices embedding $x_t$. The ablation study is conducted on the Freihand test set. The performance of different variants can be seen in Table~\ref{ablation_net}. From the table, we can clearly see that our model benefits a lot from the design of the cross-modality decoder and diffusion. Comparing the first row with the second row or the third row with the last row, error $E_{PV}$ and $E_{V}$ are largely reduced which indicates the iterative reverse process is beneficial to predict more accurate 3D hand mesh. Meanwhile, our cross-modality decoder design is also influential in reducing the prediction error from the second and last rows.

\begin{table}[!t]
    \begin{minipage}[t]{.45\linewidth}
          \caption{Ablation study on image encoder on Freihand and DexYCB($E_{PV}$/$E_{V}$(mm)$\downarrow$).}
        \centering
        \small
        \begin{tabular}{lcccc}
        \toprule
        \multirow{2}{*}{Method} & \multicolumn{2}{c}{Freihand} & \multicolumn{2}{c}{DexYCB} \\
         \cline{2-3}\cline{4-5}
        & $E_{PV}$$\downarrow$&$E_{V}$$\downarrow$&$E_{PV}$$\downarrow$&$E_V$$\downarrow$ \\
        \midrule
               HRNet-W64~\citep{hrnet} &  6.48 & 13.16 & \textbf{5.02} & \textbf{11.99}\\
               Deit-Large~\citep{deit} &  \textbf{6.38} & \textbf{12.75} & 5.18 & 12.12\\
        \bottomrule
        \end{tabular}
        \label{image_encoder}
        \end{minipage}
    \hspace*{4em}%
    \begin{minipage}[t]{.45\linewidth}
      \centering
      \small
        \caption{Effects of depth image input on Freihand and HO-3D V2($E_{PV}$/$E_{V}$(mm)$\downarrow$).}
         \begin{tabular}{lcccc}
        \toprule
        \multirow{2}{*}{Method} & \multicolumn{2}{c}{Freihand} & \multicolumn{2}{c}{HO-3D V2} \\
     \cline{2-3}\cline{4-5}
     & $E_{PV}$$\downarrow$&$E_{V}$$\downarrow$ & $E_{PV}$$\downarrow$&$E_{V}$$\downarrow$\\
           \midrule
            no depth &  6.38 & 12.75 & 9.4 & 23.9\\
            depth &  \textbf{6.30} & \textbf{12.66} & \textbf{9.3} & \textbf{23.7}\\
            \bottomrule
        \label{depth_compare}
        \end{tabular}
    \end{minipage} 
\vspace{-0.3cm}
\end{table}


\textbf{Ablation study of image encoder.} In Table~\ref{image_encoder}, we compare the performance of HRNet-W64~\citep{hrnet} and Deit-Large~\citep{deit} as image encoders. The experiments are conducted on both the Freihand and DexYCB datasets. With Deit as the image encoder, the $E_V$ and $E_J$ values are slightly lower than HRNet on Freihand. However, the trend is the opposite on DexYCB. It means means either CNN-based or transformer-based architectures can suit our cross-modality decoder. Given an effective image encoder, the performance is almost the same.

\textbf{Analysis of using extra depth image as input.} Our network is flexible that can take extra inputs into the framework. When a depth image is used as extra input to our network, the reconstruction performance can be further improved. As shown in Table~\ref{depth_compare}, depth outperforms by 0.1 mm improvement on $E_V$ in Freihand, and 0.2mm improvement on $E_V$ in HO-3D V2. With the help of a depth image, the output has a more accurate reconstruction mesh.

\section{Conclusion and Discussion}
\label{sec:concl}
In this paper, we present a novel hand mesh reconstruction method, DiffHand. Our framework is a one-stage model conditioning on hand image which progressively adds noise to clean vertices for training and reverses the process for inference. Extensive experiments show that our network design and the diffusion probabilistic model are both helpful to reach better reconstruction accuracy. For three prevailing hand mesh reconstruction datasets, we all achieve the SOTA results. 

\textbf{Broader Impacts.} In this paper, we introduce DiffHand, the first one-stage diffusion-based hand mesh reconstruction method with accurate mesh reconstruction. Our successful implementation of the diffusion model into hand mesh reconstruction encourages the application of diffusion models to more fields that have not yet used diffusion models. Diffusion models have been successfully applied to a variety of tasks, such as image denoising and super-resolution. However, these models have not yet been widely applied to hand mesh reconstruction. DiffHand demonstrates that diffusion models can be successfully applied to hand mesh reconstruction, which opens up the possibility of applying diffusion models to other 3D reconstruction tasks.

\textbf{Limitations.} Different from diffusion models in image generation tasks, for example, stable diffusion~\cite{stablediffusion} which is trained on billions of image-text pairs, our model is only trained on hundreds of thousands of image-mesh pairs. Our method may not generalize well to out-of-domain images for hand mesh reconstruction.

{
\small

\bibliographystyle{plainnat}
\bibliography{main}

}


\end{document}